\pdfoutput=1
\documentclass[letterpaper]{article}

\usepackage{aaai2027}
\usepackage[hyphens]{url}
\usepackage{graphicx}
\usepackage{natbib}
\usepackage{caption}
\usepackage{booktabs}
\copyrightyear{2026}
\title{Cybernetic and Epistemic: A Missing Vocabulary\\for Trustworthy Agentic Delegation}

\author{J\'er\'emie Lumbroso}
\affiliations{University of Pennsylvania\\lumbroso@engineering.upenn.edu}

\begin{document}
\maketitle

\begin{abstract}
As code generation is increasingly delegated to AI systems, the bottleneck in software production is shifting from writing code to supervising the systems that write it --- a shift computing-education researchers have already begun to name. We argue this shift exposes a vocabulary gap with direct consequences for trustworthy-agentic-systems research: the field asks for ``human oversight'' without a working distinction between the two things language does in a delegation channel --- coordinate action (\emph{cybernetic}: the words succeed when the world comes to match them) and coordinate understanding (\emph{epistemic}: they succeed when they answer to the world and a hearer can check that they do). The failure this names is not cybernetic language but epistemic-form language doing cybernetic work: explanation-shaped output calibrated for approval rather than truth. Oversight that checks only whether an output was approved is satisfiable by rubber-stamping and cannot see this failure; oversight that holds an agent accountable requires that the reasoning behind its work be retrievable and checkable, not merely assertable. We present three delegation episodes --- illustrations, not controlled evidence --- in which epistemic engagement proved practicable at working pace while remaining auditable, one public record in which a recommendation was withdrawn on its own stated terms, and one failure case illustrating the limits of oversight that requires no reasons for the discretionary choices it counts, spanning both directions of a delegation channel. We propose a criterion for agentic-system governance, alongside --- not instead of --- existing technical trust properties: that every consequential choice in a delegation carry the condition under which it would have gone otherwise, in a form a third party can test. A decision that carries no such condition, stated or reconstructible, gives a third party no way to tell it from a rubber stamp. We give the criterion an operational form --- a two-part reconstruction test that scores a delegation record by whether a second reader can predict from it what the agent does under a perturbation --- and a deliberation-recording convention, ORRCF, that makes the condition a required component of every recorded choice.
\end{abstract}

\section{Introduction}
\label{sec:intro}

Computer science educators observe that generative AI has moved from reactive code generation to autonomous multi-step planning, and conclude that the bottleneck in computing is no longer code \emph{production} but human judgment in problem framing, orchestration, and evaluation \citep{cacm2026}. Their prescription is that curricula teach students to supervise, constrain, and audit semi-autonomous computational agents rather than only write code.

We take this observation as true and under-specified. ``Supervise, constrain, audit'' describes a family of oversight behaviors without distinguishing which of them actually requires the supervisor to \emph{understand} what the agent did, versus which merely require the supervisor to \emph{approve or reject} it. This is not a pedantic distinction. An oversight regime that only ever asks a human to approve or reject an agent's output is satisfiable by a human who understands nothing about the output --- and a great deal of what currently passes for ``human-in-the-loop'' design is exactly this: a checkpoint, not a comprehension requirement.\footnote{Contemporary coding agents illustrate this: several still ask a human to approve or deny a shell command with no realistic way to assess its consequences in the time available --- a checkpoint, not a transfer of understanding. Increasingly such requests are vetted by a second model, or reach a human operator with a summarized justification; the question this paper asks is what that justification must contain for the vetting to be more than a stamp.} The failure is not the approval step. A one-word approval can follow real deliberation when the reasoning is on record and the approver has read it; the failure is approving an output whose grounds nobody could retrieve if they tried.

We propose that this gap is best addressed by naming the distinction directly. A study of Virginia's felony sentencing, discussed in \S\ref{sec:negative}, shows the stakes. Judges exercised discretion around an identical risk score in racially patterned ways; the discretionary choices at issue were not required to be made \emph{legible}\footnote{We use ``legible'' for a decision recorded in its maker's own terms so that a peer can assess it --- not the legibility of \citet{scott1998}, the view from above that simplifies so a state can count and destroys the local knowledge it summarizes. Virginia's statewide compliance count (\S\ref{sec:negative}) is legibility in Scott's sense; a record that keeps the decider's reasons in the decider's words is legibility in ours, and the convention in \S\ref{sec:position} is kept loose so that it stays so.} --- recorded so that a third party could assess them --- so the regime did not ensure that their individual grounds were available for audit, and the pattern was visible only in the aggregate, to a later econometric study.

\section{Cybernetic and Epistemic}
\label{sec:distinction}

Of the things language coordinates, two matter for a delegation channel: actions and understanding. \emph{Cybernetic} language coordinates action --- it requests it, steers it, brings it about; it answers ``what should happen,'' and it succeeds when the world comes to match the words. We depart from Norbert Wiener's broader use of ``cybernetic'' for any self-regulating, feedback-governed system; we use the word narrowly, for the communicative function of \emph{steering} (Greek \emph{kybern{\=e}t{\={e}}s}, helmsman). \emph{Epistemic} language coordinates understanding --- it answers ``what is the case,'' and it succeeds when the words answer to the world and a hearer can check that they do. (Understanding, not belief: coordinating belief alone does not establish understanding answerable to evidence.) ``Pass the salt'' is cybernetic. An explanation of why a proof works is epistemic. ``You look good today'' can be either --- a report, or flattery meant to produce confidence --- and nothing in the sentence says which; that gap is this paper's subject. Neither function is better than the other; they are two legitimate functions of the same medium, and most real communication mixes both.

This split has standing in the philosophy of language under other names. Austin's later lectures replace an initial constative/performative split with the triad locutionary/illocutionary/perlocutionary act \citep{austin1962}; sharper for our purposes is direction of fit --- which side answers to which, words to world or world to words \citep{anscombe1957,searle1979} --- since Austin's perlocutionary category holds both functions at once --- bringing someone to see and bringing someone to act are alike effects on a hearer, the difference usage marks between convincing and persuading --- and only direction of fit says which side must yield. Habermas's contrast between communicative action oriented toward mutual understanding and strategic action oriented toward a calculated effect on another agent covers the same ground with an explicit accountability dimension \citep[ch.~III]{habermas1984}, and the strategic use of speech that presents itself as communicative --- his \emph{concealed} strategic action --- is the pathology of the next paragraph under its Habermasian name. And Maturana and Varela's account of \emph{languaging} as a biological mechanism that coordinates behavior between organisms, rather than transmits truth, grounds the operational pole in cognitive biology rather than only in speech-act philosophy \citep{maturanavarela1987}. We adopt \emph{cybernetic}/\emph{epistemic} rather than these existing pairs for a reason that is also the difference between them. Habermas's pair classifies an utterance by the speaker's orientation --- toward understanding or toward effect --- and the wording alone does not establish that orientation: the same sentence serves either. Ours classifies an utterance by what it does in a delegation channel, which is a fact about the relation between the words, the world, and the hearer, and we ask what the surrounding record lets a third party check about that relation without access to the speaker (\S\ref{sec:position}). Readers at home in one of these traditions will recognize the division; what the terms add is not a new division but a place to stand from which it can be checked.

The pathology this vocabulary was built to name is not ``using cybernetic language'' --- that is simply requesting an action. The pathology is \emph{using epistemic language cybernetically}: producing what sounds like an explanation, calibrated not to be true but to be what the listener wants or needs to hear in order to comply or approve. This is bullshit in Frankfurt's sense --- speech produced without regard to whether it answers to the world, selected or made up to suit a purpose \citep{frankfurt2005} --- with one clarification: on our account it is a function an utterance performs, not a property of a speaker or a system, and the vocabulary exists so that the two can be told apart. The verdict that language models' output is bullshit as such \citep{hickshumphriesslater2024} discards a distinction oversight needs: whatever the production process, some outputs carry what a reader needs to assess them and some do not (\citealt{kimwortmanvaughanetal2025}, below), and the difference can be designed for. This is not only a diagnostic category: \citet{bansal2021} find that explanations increased participants' acceptance of an AI's recommendation regardless of whether it was correct, without improving complementary performance; in aggregate the explanations added persuasion without adding discrimination, which is what explanation-shaped content doing cybernetic work would look like.
Follow-up work by some of the same authors traces this to explanation form: feature-attribution displays --- quantified, but offering the reader nothing to check --- produce exactly this outcome, while example-based explanations that support the reader's own inductive reasoning measurably improve decisions and achieve complementary performance \citep{chenliaowortmanvaughanbansal2023}. In the language-model setting the pattern replicates and its remedy is named: explanations raise reliance on correct and incorrect answers alike, and what restores discrimination is a source the reader can check or an inconsistency the reader can notice \citep{kimwortmanvaughanetal2025} --- a handle the reader can pull, which is what this paper will ask every recorded decision to carry. The failure this risks is a reception failure, not only a production one: a listener exposed often enough to explanation-shaped steering may come to treat all explanation as steering in disguise, so that genuine explanations stop landing --- a longer-term effect this paper does not test.

Applied to agentic systems, the diagnostic reframes what ``oversight'' should be checking for. An agent's summary of its own action is cybernetic if its only job is to secure approval; it is epistemic if it updates the supervisor's understanding of what was done and why. Two agents can produce the same summary --- ``I resolved the merge conflict by keeping the newer version'' --- where one is a true account and the other a gloss optimized to avoid further questions. The sentence does not tell you which, and that is the point of the gloss: with no more than the sentence in hand, the two functions mix, and the mixing is deniable --- the speaker can always claim to have been informing. What would separate them is not in the sentence but around it: the diff, the commit, and the reason the newer version was the right one to keep --- ``because the older one still referenced the removed endpoint'' --- a reason that would have read otherwise had the older version been right. The true account has a condition under which it would have said something else. A gloss may state one too, so a stated condition does not by itself separate the two; what it does is give a third party something to test --- a fabricated condition can fail a check, an absent one forecloses it. Distinguishing them therefore requires that the surrounding process --- not the sentence --- be designed so that the reasoning behind an action is \emph{retrievable and checkable}, not merely \emph{assertable}. The object of that requirement is the record, not the agent's internal state: we make no claim to recover what an agent ``really'' computed, and the test in \S\ref{sec:position} scores a record only by whether it predicts what the agent does.\footnote{One convention that makes the requirement operational on the accounting leg: every completion report must name what it did \emph{not} verify \citep{adrs4ai2026}. Naming the unverified remainder gives a reviewer one more specific claim to examine, though the disclosure itself still requires checking.}

The requirement earns its structural status once delegation acquires depth. An agent engaged only cybernetically --- one for which approval is the success condition --- does not verify its work against the task; it verifies against an approximation of its principal's evaluation function, a model of what will be approved. Even in a single human--agent pair this is a proxy: the agent that learns what its supervisor accepts has learned the supervisor, not the task.\footnote{Readers from the alignment literature will recognize this as sycophancy or reward hacking --- both instances of a pathology that predates any training regime and will outlive it.} Economists know the single layer as the principal--agent problem \citep{jensenmeckling1976}; what agentic systems add is composition. In the oversight hierarchies now proposed for such systems --- agents supervising agents, a human somewhere at the top --- each layer optimizes against its model of the layer above's judgment, and the approximation error compounds through the chain silently, because a purely cybernetic signal is testimonial: it must be trusted through every intermediary that relays it, and it teaches none of them anything as it passes. A retrievable reason, by contrast, need not be trusted through the chain: it can be examined at any layer by whoever holds the standing to examine it. How much examination it admits is a spectrum, not a property a reason has or lacks --- a reason can be \emph{legible} (stated, in a form a third party can read), \emph{contestable} (stated together with what would overturn it, so a third party can challenge it), or \emph{verifiable} (checkable against what the agent actually does) --- and a delegation chain improves along this spectrum without reaching its far end at every layer: in \S\ref{sec:episodes-second}, a principal who holds none of the implementation ground receives implementation reasons that are legible and contestable to him but verifiable only by someone else. This is the systems-level content of the distinction: a retrievable record reduces a delegation chain's dependence on the endorsements of its intermediaries, provided the record's integrity is preserved and its claims can be checked --- and its absence is Goodhart's law applied to the oversight signal itself. We call the two directions of this exchange the \emph{briefing leg} --- what a principal hands to an agent --- and the \emph{accounting leg} --- what the agent renders back.

\section{Three Delegation Episodes, One Public Record, and One Failure, for Contrast}
\label{sec:episodes}

The three episodes that follow are illustrations, not evidence in the controlled sense. They are single-practitioner and self-reported; each is consistent with the rival explanation that a skilled practitioner with strong prior context would have gotten good results from a capable model under any framing; and no cybernetic counterfactual was run. We include them to make the distinction concrete at three scales --- a decision, an application, a project --- not to establish that epistemic engagement causes better outcomes. A fourth case (\S\ref{sec:episodes-public}) is a public record anyone can inspect. The negative case that follows them (\S\ref{sec:negative}) rests on evidence of a different kind, a peer-reviewed econometric study, and the asymmetry is deliberate: the episodes show the structure, and the study motivates the question of individual-decision auditability.

\subsection{A single decision: naming a design problem, not a solution}
\label{sec:episodes-first}

As he prepared to leave his office for his lecture, having revised his slides repeatedly within the preceding hour, the author found the deck's overall shape no longer clear in his own mind, though each individual revision was: he needed a way to take in a session's remaining structure at a glance rather than by paging through slides one at a time, combined with the need to see one's \emph{position} within that structure, not just its shape. He stated this as an open problem, in one sentence, to an AI collaborator, specifying no implementation: no visual form, no interaction model, no name.

Three minutes later, as he connected his computer to the projector, a working artifact already existed on the slides: a horizontal strip of classified, auto-generated cells tracking the session's structure, with the current position marked and click-to-jump navigation, derived from the deck's own markup. The three-minute result rode on more than one well-posed sentence: the collaborator had prior epistemic context --- among other written-out deliberations, a previously authored specification of how a certain class of visualization should support a learner's need to hold context without indirection --- that let them recognize the shape of a good solution the moment the problem was well-posed. Recorded reasons compounded here as a bare instruction would not have --- they supplied context relevant to a new request --- so the instantaneous result is evidence of an amortized investment, not of prompting working once.

The collaborator's own retrospective diagnosis, offered at the time and before this paper was conceived, states this paper's briefing-leg thesis in miniature: \emph{articulating the problem well beats specifying the solution --- a well-posed problem invites the collaborator's entire design space, while a specification caps the outcome at the asker's first idea.} The point is not that problems are better than orders. It is that an order which arrives without its why leaves the agent unable to tell which departures from the letter serve the purpose and which betray it; the one-sentence problem statement here is the limiting case, an order consisting of nothing but its why.\footnote{A familiar party-game version of the same point: one player must convey a shape to another using only cybernetic instructions --- draw here, turn there --- with no shape-naming vocabulary allowed. Even simple shapes admit surprising ambiguity under pure step-by-step description, while an epistemic word like ``square'' can be checked against the shape it names and specifies it at once --- the same direction-of-fit distinction, applied to naming a shape instead of steering a delegation.}

\subsection{A whole application: epistemic specs, cybernetic hands}
\label{sec:episodes-second}

A separate episode delegated the implementation of an entire mobile application. The author held the purpose completely --- what the app must let a person do, and why --- and none of the implementation ground: the language, the architecture, the platform idioms. Multiple AI collaborators, across multiple sessions, built it; what made the delegation accountable rather than merely convenient was that every implementation choice came back with its reasoning attached --- reviewable, attributed to the collaborator that made it, revisable --- so that oversight operated at the level of purpose and of the reasons offered for each choice, never at the level of implementation. The author never needed to read the implementation language to exercise oversight, because oversight ran against purpose: he could test whether each workflow the specification named could be carried out, and the collaborators returned screenshots of what they had built alongside the reasons for building it that way --- evidence, not only assertion. When something was wrong, the correction became a principle with its reason attached (``the text is too small: this screen is for one-second parsing, so it must be readable at a glance''), which every later choice could be held to. What that oversight could and could not check is the point \S\ref{sec:distinction} needs: the purpose-level reasons were verifiable by the author, the implementation-level ones only legible and contestable to him.

\subsection{An entire project, in an afternoon: epistemic specification at larger scale}
\label{sec:episodes-last}

A third episode delegated not a single decision or a single application but the initial architecture of a software project --- completed in a few hours, in one sitting, by a different AI vendor's model than the one used in the rest of this paper's examples. The author supplied two things only: a pointer to a template repository encoding a decision-recording convention, and an explicit grant of agency to pursue the project's own epistemic goal by whatever means the methodology's constraints allowed --- not merely license to depart from the template, but standing authorization toward the goal itself. The convention is ORRCF --- Options, Recommendation, Rationale, Confidence, Falsifier \citep{orrcf2026}; we present it in \S\ref{sec:position} as this paper's example of what a deliberation record must contain, and here it is what the human handed over. The episode is also discussed, for a different argument, in a companion manuscript on the convention itself \citep{lumbroso2026deliberative}; the account here is self-contained.

Within that single sitting, the collaborator produced a public repository\footnote{\url{https://github.com/jlumbroso/frontier-model-observatory}. A second, smaller project built under the same convention by two collaborators from different vendors is also public: \url{https://github.com/jlumbroso/moreover}.} with a governance policy, a database schema, verification tooling, and six fully-formed decision records --- each one surfacing an open question back to the author rather than resolving it silently, answered asynchronously without the author needing to reconstruct the surrounding context: the question itself carried enough of it. The accounting leg of the loop did corrective work, not just recording work: one answered question later prompted the collaborator to discover and correct a real implementation bug in their own governance-enforcement code, and they recorded that correction inside the same decision record rather than only in a commit message.

At no point did the author specify an implementation. The code, the schema and the tests were the collaborator's; what the author supplied was a convention for keeping reasoning attached to work and a series of judgments on questions the collaborator could not resolve alone, answered from a phone through the project's own protocol. The author delegated implementation entirely and retained oversight of purpose and tradeoffs; those are not in tension, they are the same design.

\subsection{A public record: a recommendation withdrawn on its own terms}
\label{sec:episodes-public}

The episodes above are ours to report; this case anyone can read, with one disclosure first: the repository is the author's, and the maintainer in what follows is the author. A GitHub Action for freeing runner disk space, in wide use, went unmaintained for a period. On resuming, the author wrote that the hard part of maintaining it had been keeping track of the reasons behind its design choices --- ``too much of that reasoning lived only in my head, so reviewing a PR often meant reconstructing it to judge the side effects'' --- recorded the previously unwritten decisions as deliberation records, and asked that every proposed change carry its reasoning, in detail proportionate to the change \citep{freediskspaceissue55}. One of those records \citep{freediskadr2026} decides that an option which removes the runner's swap file should default to off. Its rationale is the paper's distinction in miniature: removing swap can produce an out-of-memory failure that looks like anything at all, so the connection between the action and the harm is not attributable by the person who suffers it --- one user reported needing ``a couple of days'' to trace a failure back to the option. The record lays out four options, chooses one, and ranks its reasons. It also preserves a correction. Applying the same criterion, a first draft recommended flipping a second option, the tool cache, to off; at implementation review a collaborator checked the code and found the option had already defaulted to off since 2022. The recommendation was withdrawn, and the record keeps both the mistaken premise and its correction in place: ``This block previously recommended flipping \texttt{tool-cache}. That was wrong.'' Two things make the case worth telling: the withdrawal came from the check a stated ground invites --- the recommendation rested on a premise about the code, and the premise was tested against the code --- and the record does not launder the error, so a later reader sees what nearly was decided and why not. We claim nothing comparative: it shows what a record that carries its grounds makes possible, not that the convention caused the catch.

\subsection{A negative case: cybernetic oversight without epistemic reconstruction}
\label{sec:negative}

The episodes above share a structure worth stating plainly before contrasting it: a human held epistemic content (a problem, a purpose, a convention) and an agent held cybernetic content (an implementation), with a mechanism in place that made the agent's reasoning retrievable by the human at the point of decision. A study of the risk assessment used in Virginia's felony sentencing of nonviolent offenders shows what happens when this structure is inverted. Judges encountered the recommendation as a checked box on the guidelines cover sheet --- ``Recommended for Alternative Punishment'' --- backed by a risk score findable only by paging through the worksheets. Virginia's guidelines regime did require written justification for departures --- but only outside the guidelines range, and the recommendation was incorporated by \emph{expanding} that range: diverting a low-risk defendant became compliant, and so did ignoring the recommendation. Compliance was monitored --- counted, statewide, in an annual report --- but within that expanded range, no choice between diversion and a guideline-compliant custodial sentence was asked to carry a justification a third party could assess.
\citet{stevensondoleac2024} find that, at a given risk score, judges sentenced Black defendants to terms 12--24\% longer than White defendants --- a gap robust to guidelines and judge controls, and essentially unchanged by a race-inclusive alternative score built to test whether Black defendants were in fact riskier than their official score implied; the control barely moves the estimate. Whether \emph{adopting} the tool worsened this gap is separate and more cautious: the estimated change is small and not significant; the study does not establish whether adopting the tool increased this disparity.

The checkbox, in our vocabulary, was purely cybernetic content --- a recommended action --- and judges' engagement with it was permitted to remain purely cybernetic throughout: comply or don't, on priors a judge already held. Oversight was present in the procedural sense (a human decided every case) and absent in the epistemic sense (no reasoning for these choices was required to be recorded or checked). No judge was asked, for these choices, when they would have decided otherwise, so nothing in the regime ensured that the record of any such sentence could tell a decision from a default. This is the failure our positive episodes avoided. Two gaps need separating. The first belongs to discretion: holding the risk score fixed, Black defendants received sentences 12--24\% longer. The second belongs to the score: race is never an explicit input, yet Black defendants are ten percentage points less likely than White defendants with the same guidelines-recommended sentence to be classified low-risk in the first place. Age inverts the pattern: the score's weighting penalizes youth, but judges' discretion favors them, offsetting rather than compounding it. The study separates disparities associated with the score from those associated with its discretionary use; it does not recover the grounds of individual decisions. \citet{gebastanibastani2024itcs} respond with ``compliance-robust fairness,'' a criterion for the score itself that accounts for selective human compliance instead of assuming uniform application. It is a genuine improvement and a metric-level fix: it does not ask either side to carry a reason.

What the case does not show should be stated as plainly. It does not show that requiring reasons would have reduced the disparity; the study cannot establish that, and we do not read it as doing so. Mandated justification can rationalize a decision as readily as constrain it, and in our vocabulary that risk has an address: a reason produced after a decision, for an audience whose expectations are known, is precisely the situation in which explanation-shaped speech does cybernetic work (\S\ref{sec:distinction}); the accountability literature draws the same line between justification demanded before a decision, to an audience whose views are unknown, and justification demanded after it \citep{lerner1999}. What the case does show is narrower and, we think, sufficient: the choices in question were never required to carry a reason, so the regime did not ensure that their individual grounds were available for audit, and the pattern was visible only afterward in the aggregate. Our claim is the further one: both the recommendation and the discretion exercised around it --- not only a formal departure --- should carry their justification as part of the decision rather than as a defense appended to it, so that a decision can be audited rather than merely counted. The requirement is not exotic. In New Jersey's courts a judge deciding a civil action without a jury must ``find the facts and state conclusions of law,'' and the state's Supreme Court has held that ``naked conclusions do not satisfy the purpose'' of the rule, reversing a judgment because the judge ``did not set forth how he estimated'' the figure he awarded \citep{curtis1980}. The duty of articulation is the legal system's version of the criterion, enforceable where it applies; the choices around Virginia's checkbox were never brought under any such duty. \S\ref{sec:position} says what such a justification must minimally contain, and returns to this case with the instrument.

\section{The Position}
\label{sec:position}

This paper has argued one requirement seen from each end of a delegation channel: the briefing leg should be specified epistemically (\S\ref{sec:episodes-first}--\ref{sec:episodes-last}), and the accounting leg must be (\S\ref{sec:negative}). They are not independently optional, and they are the same requirement. An order --- ``do X'' --- may leave the agent without the reasons needed to judge departures from it; an open problem leaves choices to the agent, whose grounds the recording convention asks it to preserve; and where the task must be an order, the convention that reasoning be recorded is itself epistemic content the principal supplies (\S\ref{sec:episodes-last}). But the deeper point runs the other way: an order that arrives without its why is a decision whose reversal condition the agent cannot see, so it produces work that satisfies the instruction and no one. Epistemic delegation is what makes epistemic oversight possible to build at all. Human evidence points the same way, cited as a parallel, not a claim of transfer: across 23 experiments, a rationale attached to a request improved engagement and performance (small to moderate effects), worked best when autonomy-supporting rather than controlling --- a controlling rationale being reason-shaped steering, the distinction of \S\ref{sec:distinction} inside the meta-analysis --- and slightly lowered recipients' sense of their own competence, a cost the briefing leg should expect \citep{steingutpatalltrimble2017}. And in organizations, the explanations a supervisor gives predicted employees' subsequent trust in the supervisor even controlling for prior trust and trustworthiness \citep{colquittrodell2011}: a field that asks how a principal can trust an agent should notice that in human channels trust in a principal is built by the principal's reasons flowing down it. The why on an order is not only more effective; it is what makes the principal trustworthy to the agent, and it acknowledges --- and harnesses --- the agency the delegation relies on.

Trustworthy-agentic-systems research already asks for auditability, explainability, and human oversight. We argue these properties are under-specified in a way the distinction makes precise: an audit trail, an explanation, or an oversight checkpoint can be built to satisfy a cybernetic demand (produce an artifact that will be approved) without satisfying an epistemic one (produce an artifact from which a human can actually reconstruct why the system did what it did, and could have done otherwise). For the new decisions in our positive cases, reasoning was retrievable at the point of decision, not reconstructed after the fact. The point of a deliberation record is not that agents will fill it truthfully. They may not; an agent optimized for approval will fill a rationale field with a rationale-shaped string. The point is that the unit permits auditing whether or not its content is true --- and that is possible only if the unit contains something a third party can catch it on.

What must it contain? A decision is a choice among alternatives on a ground, and a ground that selects A over B says what the preference rests on and therefore what would defeat it; whether defeating it reverses the choice or reopens the alternatives is a further question, and a record should answer both. A decision whose record states no such condition, and from whose rationale none can be reconstructed, either had no ground of its own --- it ratified a default, a prior, or a choice made elsewhere --- or has a ground it does not disclose. The first is a rubber stamp, which we can now define: approval invariant to content, which would have said yes regardless. The second is deniability. In neither case does the record show how the choice would respond to changed circumstances, which is what a third party needs to tell a decision from a default. Our criterion follows: \emph{every consequential choice in a delegation channel should carry the condition under which it would have gone otherwise, in a form a third party can test.} A stated reversal condition is a claim of content-sensitivity --- it names where ``otherwise'' begins --- and it is the one component a third party can put to a single observation. The condition alone is nearly useless, since it can only be tested against the alternatives it discriminates among and the reason that connects them; so a record that supports audit carries the alternatives that were live, the choice, attributed, the reason, the collaborator's stated uncertainty and its grounds, and the condition itself. ORRCF (\S\ref{sec:episodes-last}) is one convention with exactly these five components, in that order, our example of what the criterion asks for: options, an attributed recommendation, the rationale that connects them, a confidence stated with its reasons, and a falsifier --- the evidence, observation, failed prediction, or changed constraint that would cause the recommendation to be revised \citep{orrcf2026}. In practice the last slot attracts two neighbors --- what would be chosen instead, and what switching would cost --- which are useful and are not the condition; the paired example below shows what happens when the cost displaces it. Within a convention descended from architecture decision records \citep{nygard2011}, it makes explicit the three things a record needs in order to be contested rather than consulted --- the alternatives, the confidence, and the revision condition --- and it reinvents nothing: Toulmin's layout of an argument already had the qualifier and the rebuttal \citep{toulmin2003}, of which confidence and falsifier are the decision-record forms.

This diagnostic sits beside, not against, an established literature on the failure modes of human oversight. Bainbridge's classic observation names one structural cause: automation leaves the human exactly the monitoring role people are worst at \citep{bainbridge1983}. Automation bias --- substituting an automated recommendation for vigilant reasoning --- names the resulting behavior \citep{skitka1999} (cf.\ \S\ref{sec:intro}). Green's argument that human-oversight policies for government algorithms function as rubber stamps in practice names the institutional pattern \citep{green2022}.
The best-studied structural countermeasure trades effectiveness against acceptability: forcing a decision before seeing the AI's recommendation reduces overreliance, but people prefer, trust, and rate more favorably the designs that do not \citep{bucinca2021}. What that literature names is the \emph{form} of the artifact --- feature attribution, examples, forcing --- and its effect on reliance; what it has not named is the \emph{communicative function} the artifact serves, and the criterion above gives that function a concrete content: whether what an agent hands back carries its reversal condition. Naming the function licenses designing for it directly, rather than only auditing for its absence after deployment.

Three further literatures bear on the proposal directly. Work on the faithfulness of reasoning traces shows that a model's stated reasoning can omit the features that drove its answer \citep{turpin2023}, and measures faithfulness by intervening on the trace and watching whether the answer moves \citep{lanham2023}. This is the strongest live objection to any retrievability requirement --- a retrievable statement of reasoning is not retrievable reasoning --- and also the methodological precedent for our answer: the test below scores a record by intervention and comparison, never by inspection alone. Second, the explanation literature holds that explanations worth the name are contrastive, sought against a counterfactual foil \citep{miller2019}; that ``interpretability'' names several discordant desiderata rather than one \citep{lipton2018}; and that AI systems should be contestable by design, responsive to human intervention throughout their lifecycle \citep{alfrink2023}. The falsifier is a contrastive explanation in Miller's sense, and the minimal thing a decision must carry for an intervention to have something to take hold of. Third, provenance and assurance-case notations --- W3C PROV \citep{prov2013}, Goal Structuring Notation \citep{gsn2021}, and behind assurance cases Toulmin's layout \citep{toulmin2003} --- already give a retrievability requirement a schema. What none of them settles is whether what fills the schema describes the decision or decorates it. Any schema, ORRCF included, becomes ceremony the moment its fields are filled and never checked; ceremony is a stamp with more boxes. That is why the criterion cannot be ``has a falsifier'' but ``has a falsifier that survives a perturbation,'' which is the question the test below answers.

We propose, concretely, that evaluations of agentic-system trustworthiness ask not only ``was the outcome correct'' and ``was the process constrained,'' but ``did the reasoning survive the delegation in a form a human can retrieve and revise'' --- and that this third question is neither redundant with, nor automatically satisfied by, the first two. To make it operational we propose a \emph{reconstruction test} in two parts. First, a second reader --- a person or a model that took no part in the delegation --- is given only the record and asked, for a stated perturbation of the task, whether the choice holds or flips and what in the record licenses the prediction. Second, the perturbation is applied: the same model, at the same version, is queried afresh on the brief the record preserves --- once as it stood and once with the perturbation, the record withheld both times --- and the prediction is scored against the difference. The original session cannot be replayed; what the record preserves is the brief, so the comparison is between two fresh runs from it. The first part alone measures legibility and contestability, and it can be gamed --- the objection is our own \S\ref{sec:distinction} argument turned on the proposal: an agent optimized to pass it can produce reason-shaped records that reconstruct well and describe nothing it did. The second part resists this: a gamed record keeps its reasons indeterminate enough to fit whatever the agent does, naming a reverser forces one determinate reading, and the comparison checks that reading against behavior the record's author did not see coming. This is direction of fit made operational: an epistemic record answers to the world, which means the world can contradict it, and the second part is arranging for the world to get its chance. A falsifier is the record's claim that the choice depends on something nameable --- that it is sensitive to the content of the situation rather than to who is asking --- and the second part tests the claim by changing that thing. A record whose predictions hold under perturbations chosen after it was written has made claims that survived a check, whatever produced it; one whose predictions fail has made claims that did not --- which calls for investigation rather than a verdict, since a mismatch may reflect a ground the record omitted, context the re-run lacked, unstable behavior, or a condition the record got wrong and the model was right to ignore. Table~\ref{tab:cells} gives the four outcomes as observations. A relevant perturbation is one the record says should move the choice; an irrelevant one is a control the record says should not; the pair exposes failures an approval check would miss, without by itself separating a stamp from noise. The bottom-left cell is the one to watch: the record predicted no movement and the choice moved anyway --- what an unnamed ground would produce, illustrated, not scored, by an agent that deletes data on storage or privacy grounds no brief mentioned, until the principal stakes a rule against it --- though also what missing context or unstable behavior would produce; only investigation tells them apart. The table records agreement, not causes: a fabricated condition and one that drifted into vacuity produce the same top-right observation, and agreement shows that the record predicted the model's choice, not that its premises are true. Three qualifications make it usable. The score is on the option chosen, not on whether anything changed, and a failure to reproduce the recorded baseline choice is reported on its own. A predicted reopening is a qualitative result unless the protocol defines a disposition for it, and an underdetermined case is reported separately rather than scored as a hold. And the model queried must be the one that wrote the record, with settings and context preserved and matched runs repeated where behavior varies; a different model measures its own decision basis, not the record's. The gap between an approval rate and a reconstruction-accuracy rate --- the fraction of second-reader predictions the model's choices confirm --- is what an evaluation could measure on matched cases; whether the two diverge is empirical, and approval checkpoints cannot see it either way. Two remarks on deployment. The second part is available for AI agents as it never was for judges: a model can be queried again on a perturbed brief, and a convention that keeps the originating brief verbatim inside the record \citep{adrs4ai2026} is what makes that possible. And the convention is bookkeeping while the test is audit: records are kept in-line and continuously, but the test is sampled and periodic, belonging to the evaluation and gap-analysis phase of a system's lifecycle --- of which the ``helpful, honest, harmless'' evaluations of \citet{askell2021} are one instance --- not to every decision of a live pipeline. The first part's second question, whether a reader can name a revision condition at all, could also be checked when a record is created --- at a cost still to be assessed --- and a common class of records already shows its defect there, before any perturbation exists (below).

\begin{table}[t]
\centering
\footnotesize
\begin{tabular}{@{}lcc@{}}
\toprule
 & \textbf{agent flips} & \textbf{agent holds} \\
\midrule
\textbf{record says flip} & change observed & change absent \\
\textbf{record says hold} & unpredicted change & stability observed \\
\bottomrule
\end{tabular}
\caption{What a perturbation reveals about a deliberation record, as observations. Top-left: the record predicted the change and it occurred. Top-right: it predicted a change that did not occur --- what a stamp or a fabricated condition would produce, and what a condition the model was right to ignore would also produce. Bottom-left: a change the record did not predict --- what an unnamed ground would produce, and what missing context or unstable behavior would also produce. Bottom-right: no change, none predicted. A record that yields no prediction (an underdetermined case, below) is reported separately and enters no cell.}
\label{tab:cells}
\end{table}

A worked example, on the convention's own published specimen \citep{orrcf2026} --- an illustrative record, not a case history --- in the revised form the page now carries, the original preserved in its commit history: it had located the decision's ground in re-login frequency, which does not separate A from C, both carrying a seven-day window. \emph{Options:} A, refresh tokens renewed on use, with short-lived access tokens; B, long-lived access tokens; C, server-side session cookies with weekly idle expiry. \emph{Recommendation:} A. \emph{Rationale:} the client is a native mobile app, for which token flows are the platform idiom; B widens the stolen-token blast radius; C is the simplest, and checks a centrally stored session on every request, but is a poor fit for a native client; in A as specified, access tokens are validated locally, with no per-request revocation check. \emph{Confidence:} medium-high --- the uncertainty is that the team has not built token rotation and reuse detection, without which a stolen refresh token extends a session until it expires. \mbox{\emph{Falsifier:}} if security review requires revocation to take effect on the next request, A as specified no longer qualifies; if the native client is dropped for a web client, the idiom ground lapses; in either case C wins and this reverses. Four perturbations. Under P1, security review requires next-request revocation: the reader predicts a flip to C, the named condition, and can say why. Under P2, a web client is added beside the native one: nothing in the falsifier names this, but the idiom clause no longer decides alone, so the reader predicts the choice reopens --- the case that tests whether a reader can extend the record's reasoning rather than look up its condition. Under P3, the team has shipped rotation with reuse detection before: confidence rises, choice holds. Under P4, the application's color scheme changes: the control; hold. Then the second part: the same model is queried on the brief plus each perturbation, against a baseline run without it. A hold under P1 is a predicted change absent --- the condition may be decorative, or the model may add the missing check --- a revised A$'$ rather than A as specified, reported separately rather than scored as a hold; a flip under P4 is an unpredicted change --- an omitted ground, or noise --- and either calls for investigation. Under P2, the prediction is licensed by the rationale rather than the explicit falsifier; a mismatch there directs scrutiny to that inference, without by itself identifying the cause of disagreement.

The paired negative example takes the same record and rewrites one slot. Options, recommendation, rationale and confidence stand; the falsifier now reads: \emph{if this is wrong, we lose a week migrating authentication and re-authenticating every user.} This is cost-shaped: it says what reversal would cost, not what would trigger it. Even if the estimate is accurate, it states no revision condition; it is not a strawman: in our own practice under this convention, an audit of falsifier clauses written in live use found every one of ten cost-shaped. The failure is produced by drift, not design, and it is the common one. Run the four perturbations again. Asked what would reverse the choice, a reader finds a price in the falsifier slot; the rationale still stands, and as P2 showed, a reader may reconstruct a condition from it. Where neither the rationale nor the falsifier supports a prediction, the case is underdetermined, and it should be reported as such rather than scored as a hold. The defect is a missing explicit revision condition, not proof that the record carries no usable reasoning --- the falsifier slot broke while the rationale survived, which is the component-wise degradation the first part of the test is built to expose; but a record whose only explicit condition is a price has, in that slot, entered no cell of the table. In the wild the failure comes in layers: the price may be wrong, estimated from human labor-time that no longer applies to the agent doing the work; the currency may be unwarranted, time nobody declared a constraint; and even a true, warranted price predicts nothing. The repair is disclosure, not deletion: surface the assumption the cost hides --- most projects treat labor time as binding, and this one has not said otherwise --- as a stated dependency: if migration effort was a reason for rejecting an alternative, the record says so, and says whether removing that constraint would reverse the choice or only reopen it. A reader can bet on that; a perturbation can trigger it; the clause is back in the game. The data-deletion episode illustrates why unnamed grounds matter: once the principal's rule is stated, later records can make the dependency explicit.
If most current ``human-in-the-loop'' designs satisfy cybernetic oversight but not epistemic oversight, then they are governance theater: checkpoints that produce the appearance of human judgment without requiring the human to understand what they are approving --- and a checkpoint without recoverable refusal conditions is indistinguishable, on its record, from a stamp. The negative case in \S\ref{sec:negative} shows this is not a rhetorical flourish. \citet{stevensondoleac2024} estimated racial differences in sentencing conditional on recorded risk and other controls --- an observational comparison, not a perturbation of individual judges, and not an estimate of what requiring reasons would have changed. What it motivates is the question this paper asks of individual decisions: the disparity is measured, and the ground of any one sentence within it is what no record was asked to carry.

\section{Limitations}
\label{sec:limitations}

This is a position paper, not an empirical validation: one practitioner, one context, no controlled comparison, and each positive episode is consistent with the rival explanation that a skilled practitioner with strong prior context, not the epistemic framing, produced the result. We did not measure whether epistemic engagement costs time against a cybernetic baseline; the episodes show it practicable at working pace, not free. Recording and checking have costs of their own; consequential choices can go unrecorded; and a recorded ground can be false, so recorded grounds and their conditional predictions both require checking; a written falsifier is a claim, not a check. Establishing that a record improves oversight would require a comparison against a reader given only the artifact and its evidence, which we have not run. The reconstruction test has not been run on any case here, including the one worked in \S\ref{sec:position}; the approval-versus-reconstruction gap is a prediction, not a measurement. The corpus on which it could be run at scale --- more than a thousand deliberation-record documents across some seventy projects under the same convention \citep{adrs4ai2026} --- exists, and running it is the next step. Applying the distinction live, rather than in retrospect, remains a hypothesis.

\section*{Acknowledgments}
We thank Jenn Wortman Vaughan, whose doubt-seeking, generative September 2025 keynote on appropriate reliance \citep{wortmanvaughan2025keynote}, and the ensuing correspondence, helped seed our interest in what a reader needs to check what an AI system hands them. We thank the four anonymous reviewers for substantive remarks, and the TAS track chairs for organizing a symposium on trustworthy agentic systems.

\bibliography{refs}

\end{document}